\documentclass{article}

\PassOptionsToPackage{numbers,square,comma,sort}{natbib}
\usepackage[preprint]{neurips_2024}

\usepackage[utf8]{inputenc} 
\usepackage[T1]{fontenc}    
\usepackage{url}            
\usepackage{booktabs}       
\usepackage{amsfonts}       
\usepackage{nicefrac}       
\usepackage{microtype}      
\usepackage{xcolor}         
\usepackage{amssymb}
\usepackage{float}
\usepackage{amsmath}
\usepackage{amsthm}
\usepackage{graphicx}
\usepackage{diagbox}
\usepackage{tabularx}
\usepackage{bm}
\usepackage{bbm}
\usepackage{algorithm}
\usepackage{algorithmic}
\usepackage{setspace}
\usepackage{subfigure}

\usepackage{amsfonts}
\usepackage{float}
\usepackage{enumitem,graphicx}
\usepackage{wrapfig}
\usepackage{multirow}
\usepackage{xcolor}
\usepackage[pagebackref=true,breaklinks=true,colorlinks,bookmarks=false]{hyperref}

\def\red#1{\textcolor{red}{#1}}

\def\x{\bm{x}}

\def\vtheta{\bm{\theta}}

\def\gL{\mathcal{L}}

\makeatletter
\DeclareRobustCommand\onedot{\futurelet\@let@token\@onedot}
\def\@onedot{\ifx\@let@token.\else.\null\fi\xspace}

\def\ie{\emph{i.e.}}

\def\etal{\emph{et al. }}

\long\def\comment#1{}
\newcommand*{\email}[1]{%
    \normalsize\href{mailto:#1}{#1}\par
    }

\usepackage{times}
\usepackage{epsfig}
\usepackage{graphicx}
\usepackage{amsmath}
\usepackage{amssymb}
\usepackage{dsfont}
\usepackage{xcolor}

\def\vtheta{{\bm{\theta}}}

\def\gL{{\mathcal{L}}}

\def\gX{{\mathcal{X}}}
\def\gY{{\mathcal{Y}}}

\DeclareMathOperator*{\argmin}{arg\,min}

\newcommand{\RNum}[1]{\MakeUppercase{\romannumeral #1}}

\title{Breaking the False Sense of Security in Backdoor Defense through Re-Activation Attack}

\author{
Mingli Zhu\textsuperscript{1}\ \ \ \ 
Siyuan Liang\textsuperscript{2} \ \ \ \ 
Baoyuan Wu\textsuperscript{1}\thanks{Corresponds to Baoyuan Wu (\email{wubaoyuan@cuhk.edu.cn}).} \\
\textsuperscript{1}School of Data Science, \\
The Chinese University of Hong Kong, Shenzhen, Guangdong, 518172, P.R. China \\
\textsuperscript{2}National University of Singapore, Singapore
}

\begin{document}

\maketitle

\begin{abstract}

Deep neural networks face persistent challenges in defending against backdoor attacks, leading to an ongoing battle between attacks and defenses.  
While existing backdoor defense strategies have shown promising performance on reducing attack success rates, can we confidently claim that the backdoor threat has truly been eliminated from the model?  
To address it, we re-investigate the characteristics of the backdoored models after defense (denoted as defense models). 
Surprisingly, we find that the original backdoors still exist in defense models derived from existing post-training defense strategies, and the backdoor existence is measured by a novel metric called \textit{backdoor existence coefficient}. 
It implies that the backdoors just lie dormant rather than being eliminated.
To further verify this finding, we empirically show that these dormant backdoors can be easily re-activated during inference, by manipulating the original trigger with well-designed tiny perturbation using universal adversarial attack. 
More practically, we extend our backdoor re-activation to black-box scenario, where the defense model can only be queried by the adversary during inference, and develop two effective methods, \ie, query-based and transfer-based backdoor re-activation attacks. 
The effectiveness of the proposed methods are verified on both image classification and multimodal contrastive learning (\ie, CLIP) tasks.
In conclusion, this work uncovers a critical vulnerability that has never been explored in existing defense strategies, emphasizing the urgency of designing more robust and advanced backdoor defense mechanisms in the future.

\end{abstract}

\section{Introduction\label{sec1}}

The pervasive application of Deep Neural Networks (DNNs) across safety-critical domains like facial recognition and autonomous driving \cite{kaur2020facial,liu2020computing} has underlined their significance and profound impact in industrial and academic spheres. Despite their transformative potential, DNNs are known to be vulnerable to malicious threats \cite{biggio2018wild,kumar2020adversarial,Liang_2021_ICCV,liang2022large}, which compromise the integrity and reliability of systems. One of the representative threats is backdoor attacks \cite{gu2019badnets,li2021invisible}, where an adversary pre-defines a "trigger" and embeds it within limited training data such that the backdoored model will misclassify trigger-containing inputs into specific target categories while appropriately processing benign inputs. 

A successful backdoor attack consists of two stages: (1) the embedding of the backdoor within the model during training; and (2) its subsequent activation during inference stage \cite{wu2023attacks}. To identify \cite{dong2021black} and mitigate the harmful impacts of backdoor attacks, substantial efforts have been made  ranging from dataset segmentation \cite{chen2019detecting,tran2018spectral}, trigger inversion \cite{wang2019neural,wangunicorn}, model pruning \cite{zheng2022data,wu2021adversarial}, and fine-tuning based defenses \cite{li2021neural,zeng2022adversarial}. 
While these existing defense mechanisms aim at decreasing the attack success rates (ASR) \cite{wubackdoorbench} of corresponding backdoored models, a fundamental question arises: \textit{can we confidently claim that the backdoor threat has truly been eliminated from the model? } In this work, we use the term \textbf{defense model}(s) to denote those models which have initially been poisoned to backdoored models and subsequently defended using some defensive techniques, for convenience.

\begin{wrapfigure}{r}{0.39\textwidth}
\vspace{-0.6em}
\centering
\includegraphics[width=0.4\textwidth]{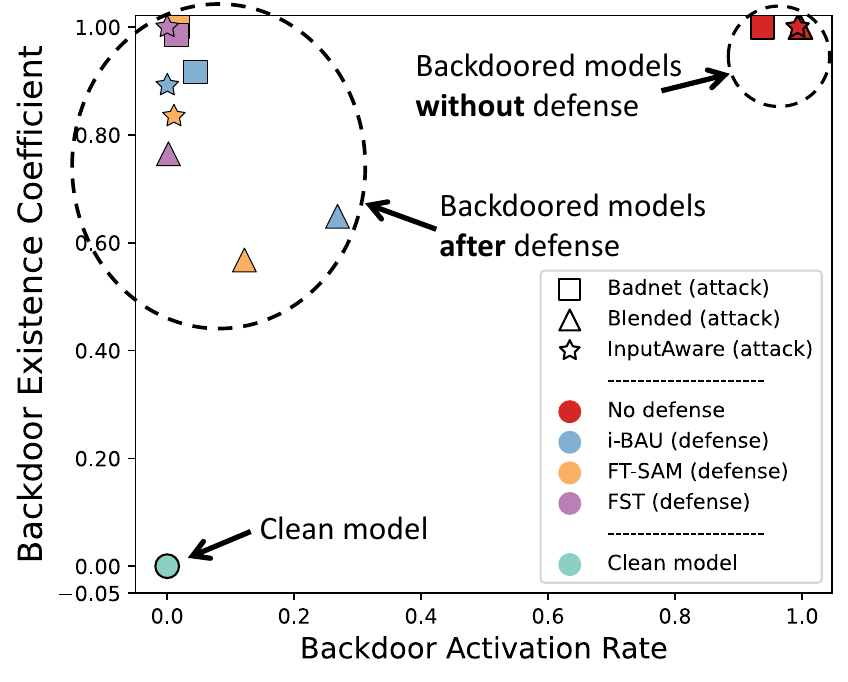}
 \vspace{-2.1em}
\caption{Comparative analysis of backdoor existence coefficient and backdoor activation rate across different models.}
\vspace{-0.15in}
\label{fig:moti}
\end{wrapfigure}
To answer above question, we introduce an innovative concept \textit{backdoor existence coefficient} (\textbf{BEC}) to quantify the extent of backdoor presence within models. 
Using BEC, we can re-investigate the backdoor existence in existing defense models \cite{li2021neural,zeng2022adversarial,wei2024shared}.
Specifically, the BEC measures the similarity of activation among backdoor-related neurons in the poisoned samples between the backdoored model and its corresponding defense model.
Fig. \ref{fig:moti} presents the relationship between BEC and backdoor activation (indicated by ASR) across three different attack and defense methods for comparison. In this figure, distinct shapes and colors denote various attack and defense methods, respectively. As depicted in the figure, even though the ASRs decline nearly to zero which implies that defense models perform comparably to clean models, the BECs in the defense models remain significantly high. This notable observation implies that the original backdoors just lie dormant rather than being eliminated in defense models.

Inspired by above observations, we pose a question: Since the original trigger fails to activate the original backdoor, is it possible to unearth a variant of the original trigger that is capable of re-activating the backdoor?
Given that in real-world scenarios where the adversary cannot modify the defense model, our objective is to modify the original trigger, thereby facilitating backdoor re-activation in defense models during inference stage. To verify this feasibility, we formulate the backdoor re-activation task as constrained optimization problem with the goal of searching for a minimal universal adversarial perturbation on the original trigger. Consequently, this general technique can be seamlessly combined with any prevailing backdoor attacks to re-activate backdoor effect in defense models in their inference time. 
To demonstrate the real-world threat posed by backdoor re-activation attack, we also expand our method to black-box and transfer attack scenarios, where adversaries are limited to querying the model without access to its internal mechanisms. Nowadays, multimodal contrastive learning (MMCL) has impressed us with its performance across a range of tasks and backdoor threats in MMCL have also been broadly studied. In this work, we consider both image classification and multimodal tasks, demonstrating the universality and adaptability of our approach. Extensive experimental results on nine different attacks and eight state-of-the-art defenses across four benchmark datasets and three model architectures demonstrate the effectiveness of our method. Our work reveals a new vulnerability in existing defense strategies, emphasizing the need for more robust and advanced defense mechanisms in the future.

Our main contributions are threefold: 
\textbf{1)} We re-investigate existing defense methods, and reveal that the original backdoor still exists in the model even after defense, though it cannot be activated by the original trigger. 
\textbf{2)} We develop a novel optimization problem to re-activate the original backdoor during inference by perturbing the original trigger, under white-box, black-box, and transfer attack scenarios.  
\textbf{3)} We demonstrate the effectiveness of the proposed method with extensive experiments on both image classification and the emerging multi-modal contrastive learning tasks. 
\vspace{-2mm}

\section{Related work}
\paragraph{Backdoor attacks.}

Backdoor attacks \cite{wubackdoorbench,jiang2023color,liu2020reflection,gao2023backdoor,wang2023robust} are a significant security threat in DNNs. As summarized by Wu \etal \cite{wu2023attacks,wubackdoorbench}, a successful backdoor attack consists of two components: \textit{backdoor injection} during pre-training or training stage, and \textit{backdoor activation} during inference stage. Backdoor injection could be divided into data poisoning attack at pre-training stage and training-controllable attack at training stage.
During a data poisoning attack, an adversary releases a poisoned dataset to plant backdoors. Representative works include BadNets \cite{gu2019badnets}, Blended \cite{chen2017targeted}, LF \cite{zeng2021rethinking}, SSBA \cite{li2021invisible}, and Trojan \cite{Trojannn}.
For training-controllable attack \cite{zhao2022defeat}, an adversary takes control of the training process to optimize triggers and inject backdoors. Notable examples are Input-Aware \cite{nguyen2020input} and WaNet \cite{nguyen2021wanet}. In inference stage, the adversary uses the poisoned samples to activate backdoors in the backdoored model, thereby achieving a successful attack.

While backdoor attacks are prevalent in supervised learning, backdoor threats also exist in domain of multi-modal contrastive learning (MMCL). Carlini \etal  
 \cite{carlini2022poisoning} are the pioneers to unveil backdoor threats in MMCL, demonstrating that as few as $0.0001\%$ of images can trigger a successful attack. More recently, sophisticated approaches have been introduced  \cite{bai2023badclip}. For instance, TrojanVQA \cite{walmer2022dual} is designed for the multi-modal visual question answering task, while BadCLIP \cite{liang2023badclip} shows that their attack can persist in effectiveness against backdoor defenses.

While a variety of attack methods have been proposed, they primarily focus on enhancing attack success rate during backdoor injection stage and employ the same trigger to activate backdoors in inference stage. They did not consider that the model might be fine-tuned or defended by users, and the original triggers fail to activate backdoors in inference stage. Although Qi \etal \cite{qi2022revisiting} attempted to enhance backdoor signal during inference stage, they did not consider defensive techniques in depth, and their attack lacks universality. In this work, we focus on a general backdoor attack method during inference time, researching on how to re-activate the dormant backdoors in defense models.

\paragraph{Backdoor defenses.}

A range of works \cite{khaddaj2023rethinking,li2021anti,huang2022backdoor,min2024towards,zhang2022purifier} focusing on backdoor defenses have been put forward to address the threat of backdoor attacks. Considering the defense stages, four main categories emerge: pre-processing defenses, training-stage defenses, post-training defenses, and inference stage defenses \cite{wu2024backdoorbench,wu2023defenses}. 
Pre-processing defenses \cite{chen2019detecting,zhu2023vdc,khaddaj2023rethinking,yang2024not} aim to filter out poisoned samples from poisoned dataset. 
Training-stage strategies \cite{li2021anti,huang2022backdoor,chen2022effective} consider that the defender has access to both training samples and the model, and mitigates backdoor effects during training process. They leverage discrepancies between poisoned and benign samples to filter out suspicious instances. 
Post-training defenses \cite{min2024towards,zhang2022purifier,chen2022linkbreaker,wang2023mm,zhu2024neural} focus on removing backdoor effect from backdoored models through pruning potential backdoor neurons \cite{wu2021adversarial}, backdoor triggers reversion and unlearning \cite{wang2019neural}, or enhancing fine-tuning processes for backdoor mitigation \cite{zhu2023enhancing}. 
Inference stage defenses aim at preventing backdoor activation with samples detection or samples recovery techniques \cite{zhu2023vdc}.
In the domain of MMCL, CleanCLIP \cite{bansal2023cleanclip} is the first to defend the MMCL model using MMCL loss and self-supervised learning within each modality with clean samples. Additionally, RoCLIP \cite{yang2024robust} introduces a robust pre-training approach, which focuses on disrupting the link between poisoned image-caption pairs. In this work, we focus on backdoor re-activation attack and thus mainly consider our attack against post-training backdoor defenses.

\section{Methodology\label{sec3}}
In this section, we introduce our threat model and methods for image classification task for clarity. For the formulation and methods for multimodal contrastive learning, please refer to \textbf{Appendix} \red{A}.

\subsection{Threat model\label{sec3.1}}
\paragraph{Notations.}
For the image classification task, the training dataset is $\mathcal{D}=\{(\x^{(i)},y^{(i)})\}_{i=1}^{n}  \subseteq \gX \times \gY$, where $\gX \subset \mathbb{R}^d$ and $\gY = \{1,\dots,K\}$ are input space and label set, respectively. Given an input $\x$, we define a deep neural network with $L$ layers as:
\begin{equation}
    f(\x) = f^{(L)} \circ f^{(L-1)} \circ \cdots \circ f^{(1)}(\x),
\end{equation}
where $f^{(l)}$ is the function in the $l^{th}$ layer of the network, $1 \leq l \leq L$. The feature map of the $l^{th}$ layer is denoted as $m^{(l)}(\x) \in \mathbb{R}^{c_l\times h_l \times w_l}$, and $f_k(\x)$ represents the logit of the $k^{\text {th}}$ class.

Before introducing our methods, we first outline the pipeline of backdoor attack and defense. As summarized in \cite{wu2023attacks} and shown in Tab. \ref{table0}, the whole pipeline of backdoor attack and defense involves four stages:
\begin{enumerate}[leftmargin=8mm,label=\Roman*.]
    \item \textbf{Pre-training stage}: An adversary conducts data poisoning backdoor attack, which involves revising a small fraction of $\mathcal{D}$ to generate poisoned dataset $\mathcal{D}_p=\{(\x^{(i)}_{\bm{\xi}},t)\}_{i=1}^{n_p}$ by injecting a trigger $\bm{\xi}$ into the image and changing the corresponding label into target label $t$. 
    \item \textbf{Training stage}: An adversary controls the training process to inject backdoors into model $f_{\vtheta_\text{A}}$.
    \item \textbf{Post-training stage}: A defender receives the poisoned model, and can gather some benign samples to remove backdoor effect from the model, defined as $f_{\vtheta_\text{D}}$.
    \item \textbf{Inference stage}: With the defense model $f_{\vtheta_\text{D}}$, the original trigger fails to activate the backdoor, \ie, $f_{\vtheta_\text{D}}(\x_{\bm{\xi}})\neq t$. The goal is to re-activate backdoors, \ie, $f_{\vtheta_\text{D}}(\x_{\bm{\xi}^{\prime}}) = t$, where $\bm{\xi}^{\prime}=\bm{\xi}+\Delta_{\bm{\xi}}$.
\end{enumerate}

Existing backdoor attacks primarily focus on achieving high attack success rates (ASR) in backdoor injection stages (\RNum{1} and \RNum{2}), with little consideration for the defensive impact in stage \RNum{3}. \textit{Given the failures of $\x_{\bm{\xi}}$ in attacking $f_{\vtheta_\text{D}}$, our work focuses on the backdoor re-activation attack in stage \RNum{4}.} 

\begin{table}[t]
  \caption{Illustration of the pipeline of backdoor attack and defense. }
  \label{table0}
  \centering
    \scalebox{0.91}{
  \begin{tabular}{llll}
    \toprule
    \cmidrule(r){1-4}
    Stage     & Task description & Input/Output    & Goal \\
    \midrule
      Reference    &   Clean model training   & $\mathcal{D}$/$f_{\vtheta_\text{C}}$ &
      $f_{\vtheta_\text{C}}(\x)=y$, $f_{\vtheta_\text{C}}(\x_{\bm{\xi}})\neq t$ \\
    \RNum{1}: Pre-training \& \RNum{2}: Training   &   Backdoor injection & $\mathcal{D}$/$f_{\vtheta_\text{A}},\mathcal{D}_p$ &  $f_{\vtheta_\text{A}}(\x)=y$, $f_{\vtheta_\text{A}}(\x_{\bm{\xi}})= t$ \\
    \RNum{3}: Post-training  & Backdoor defense  & 
    $f_{\vtheta_\text{A}}$/$f_{\vtheta_\text{D}}$ &
    $f_{\vtheta_\text{D}}(\x)=y$, $f_{\vtheta_\text{D}}(\x_{\bm{\xi}})\neq t$ \\
    \RNum{4}: Inference    & Backdoor re-activation  & 
    $\x, \bm{\xi}, f_{\vtheta_\text{D}}$/$f_{\vtheta_\text{D}}(\x_{\bm{\xi}^{\prime}})$ 
    & 
    $f_{\vtheta_\text{D}}(\x)=y$, $f_{\vtheta_\text{D}}(\x_{\bm{\xi}^{\prime}}) = t$ \\
    \bottomrule
  \end{tabular}}
\end{table}

\subsection{Backdoor existence coefficient\label{sec3.2}}
While the model performance in Tab. \ref{table0} suggests that $f_{\vtheta_\text{D}}$ and $f_{\vtheta_\text{C}}$ are analogous, we argue that in terms of the backdoor effect, $f_{\vtheta_\text{D}}$ and $f_{\vtheta_\text{A}}$ are actually more closely aligned, which indicates the persistent existence of backdoor in model $f_{\vtheta_\text{D}}$. To verify this, we need a metric to measure the quantity of backdoor existence within a model. An effective indicator should be capable of quantifying the similarity of backdoor effect between backdoored model $f_{\vtheta_\text{A}}$ and the target defense model $f_{\vtheta_\text{D}}$ across the entire models. To achieve this, we propose a new metric, \textit{Backdoor Existence Coefficient} (BEC), which is calculated through the following three steps:
\begin{enumerate}[leftmargin=8mm]
    \item \textbf{Backdoor neuron identification}: Firstly, we need to identify backdoor-related neurons. Zheng \etal \cite{zheng2022data} proposed Trigger-activated Change (TAC) to quantify the correlation between backdoor impact and neurons (see \textbf{Appendix} \red{C} for details).
    With this metric, backdoor-related neurons in $f_{\vtheta_\text{A}}$ are identified for each layer. Thus, the feature maps corresponding to these neuron indices are selected for each model, denoted as $\Tilde{m}^{(l)}_{\text{A}}(\x_{\bm{\xi}})$, $\Tilde{m}^{(l)}_{\text{D}}(\x_{\bm{\xi}})$, and $\Tilde{m}^{(l)}_{\text{C}}(\x_{\bm{\xi}})$, respectively. Denote the feature maps across dataset $\mathcal{D}_p$ as $\Tilde{m}^{(l)}(\mathcal{D}_p)\in \mathbb{R}^{n_p \times (\Tilde{c}_l\times h_l \times w_l)}$.
    \item \textbf{Backdoor effect similarity metric}: In order to measure the backdoor effect similarity between models, we employ Centered Kernel Alignment (CKA) \cite{kornblith2019similarity} (see \textbf{Appendix} \red{C} for details) to quantify the similarity between these matrices. The similarity in backdoor effects between $f_{\vtheta_\text{D}}$ and $f_{\vtheta_\text{A}}$, calculated through the use of corresponding features, can be computed as:
    \begin{equation}
    S_{\text{D},\text{A}}^{(l)}(\mathcal{D}_p) = \text{CKA}\left(\Tilde{m}_{\text{D}}^{(l)}(\mathcal{D}_p), \Tilde{m}_{\text{A}}^{(l)}(\mathcal{D}_p)\right),
    \end{equation}
     and $S_{\text{C},\text{A}}^{(l)}(\mathcal{D}_p)$ is computed accordingly.
     \item \textbf{Backdoor existence coefficient computation}: The BEC is the average of normalized backdoor effect similarity across all layers. By assigning the BEC of $f_{\vtheta_\text{A}}$ a value of 1 and $f_{\vtheta_\text{C}}$ a value of 0, the computation can proceed as follows:
     \begin{equation}
    \rho_{\text{BEC}} (f_{\vtheta_{\text{D}}},f_{\vtheta_{\text{A}}},f_{\vtheta_{\text{C}}};\mathcal{D}_p) = \frac{1}{N}\sum_{l=1}^{N} \frac{S_{\text{D},\text{A}}^{(l)}(\mathcal{D}_{p})-S_{\text{C},\text{A}}^{(l)}(\mathcal{D}_{p})}{S_{\text{A},\text{A}}^{(l)}(\mathcal{D}_{p})-S_{\text{C},\text{A}}^{(l)}(\mathcal{D}_{p})}\in [0,1].
\end{equation}
\end{enumerate}

\textbf{Remark.} Denote $\rho_{\text{BEC}} (f_{\vtheta_{\text{D}}},f_{\vtheta_{\text{A}}},f_{\vtheta_{\text{C}}};\mathcal{D}_p)$ as $\rho_{\text{BEC}} (f_{\vtheta_{\text{D}}})$ for simplicity. The higher the value $\rho_{\text{BEC}}(f_{\vtheta_{\text{D}}})$, the greater the existence of backdoors in the model. We utilize $\text{BEC}$ to signify backdoor existence and employ ASR to quantify the extent of backdoor activation. As shown in Fig. \ref{fig:moti}, the BEC remains consistently high across various defenses, despite backdoor activation being low.

\subsection{Backdoor re-activation attack\label{sec3.3}}

Motivated by the fact analyzed above that the original backdoor still exists in the defense model $f_{\boldsymbol{\theta}_\text{D}}$, here we explore the possibility to re-activate the backdoor during inference. Since the adversary cannot modify $f_{\boldsymbol{\theta}_\text{D}}$ during inference, one feasible solution is to modify the original trigger $\bm{\xi}$. Specifically, we propose to pursue a new trigger $\bm{\xi}'$ by perturbing $\bm{\xi}$, \ie, $\bm{\xi}' = \bm{\xi} + \Delta_{\bm{\xi}}$, such that $\bm{\xi}'$ could re-activate the original backdoor, \ie, $f_{\boldsymbol{\theta}_\text{D}}(\x_{\bm{\xi}'})=t$. 
In the following, we will present how to obtain a successful trigger perturbation $\Delta_{\bm{\xi}}$ under white-box, black-box, and transfer attack scenarios, respectively.

\paragraph{White-box backdoor re-activation attack.} 
In white-box scenario, the adversary has access to the parameters of $f$ but cannot manipulate them. In this case, we could obtain $\Delta_{\bm{\xi}}$ by solving the following constrained optimization problem: 

\begin{equation}\label{eq5}
\min_{\|\Delta_{\bm{\xi}}\|_p\leq \rho} \gL_{tot}(\Delta_{\bm{\xi}};\mathcal{D}_{p}, f)=\sum_{(\x_{\bm{\xi}},t) \in \mathcal{D}_{p}}  \gL_{\text{CE}}(f(\x_{\bm{\xi}+\Delta_{\bm{\xi}}}), t)-\lambda \log\left(1-\max_{k\neq t} \frac{e^{f_{k}(\x_{\bm{\xi}+\Delta_{\bm{\xi}}})}}{\sum_{i=1}^{N} e^{f_{i}(\x_{\bm{\xi}+\Delta_{\bm{\xi}}})}}\right),
\end{equation}
where $\|\cdot\|_p$ means $\ell_p$ norm, $\rho$ is the perturbation bound, $\gL_{\text{CE}}$ is cross-entropy loss, and $\lambda>0$ is a hyper-parameter. This problem can be easily solved using project gradient descent (PGD) \cite{madry2017towards}.

\paragraph{Black-box backdoor re-activation attack.}
Although the re-activation attack under the white-box scenario is easy to implement, it may be impractical. 
Thus, we also consider the practical black-box scenario, where the adversary lacks information to the defense model and can only query the model and obtain the predicted score. 
Consequently, the above problem (\ref{eq5}) is no longer directly optimized by the PGD algorithm.  

Inspired by existing black-box adversarial attacks \cite{andriushchenko2020square,chen2020rays}, we propose a novel random search based optimization algorithm. Specifically, we extend the query-based  black-box adversarial attack method Square Attack \cite{andriushchenko2020square} that was designed for optimizing sample-specific perturbation, to solve problem (\ref{eq5}), dubbed Universal Square Attack. 
Its overall procedure is summarized in Alg. \ref{alg2}.

\paragraph{Transfer-based backdoor re-activation attack.} 
In addition to the query-based black-box attack, we also explore the transfer scenario where the adversary lacks prior knowledge of the target defense model and has restricted query limits. Consequently, leveraging transfer attacks becomes a viable strategy for attacking.
The main idea is that the adversary can imitate defense process to get some defense models themselves. Then these defense models can serve as surrogate models to generate perturbation $\Delta_{\bm{\xi}}$ as follows:
\begin{equation}\label{eq6}
\Delta_{\bm{\xi}}^{*} = \argmin_{\|\Delta_{\bm{\xi}}\|_p\leq \rho}\sum_{i=1}^{M} \gL_{tot}(\Delta_{\bm{\xi}};\mathcal{D}_{p}, f_{i}).
\end{equation}

Overall, we propose a universal backdoor re-activation attack that aims to enhance the performance of existing backdoor attack methods during inference. We have explored three scenarios—white-box attack (WBA), query-based black-box attack (BBA), and transfer attack (TA). 

Besides, we would like to emphasize again that the proposed attack can be naturally extended to multi-modal learning task, other than the classification task demonstrated above.  The details are presented in \textbf{Appendix} \red{A}.

\begin{algorithm}[t]
\caption{Black-box Backdoor Re-Activation Attack via Universal Square Attack (BBA) \cite{andriushchenko2020square}}\label{alg2} 
\begin{algorithmic}[1]
\STATE \textbf{Input:} Defense model $f$, training dataset $\mathcal{D}_p$, image shape $c,h,w$, norm $p$, perturbation bound $\rho$, target label $t\in {1,\dots,K}$, number of iterations $N$, termination condition $\epsilon$.
\STATE \textbf{Output:} Perturbation $\Delta_{\bm{\xi}}^{*}$ as in Eq. \ref{eq5}.
\STATE $\hat{\x} \leftarrow \x + \text{init}(\Delta_{\bm{\xi}})$ for $\x \in \mathcal{D}_p$, \quad $l^{*} \leftarrow \mathcal{L}_{tot}(\mathcal{D}_{p},\Delta_{\bm{\xi}})$.
\FOR {$i=0,...,N -1$}
\STATE \textbf{if} ASR $>1-\epsilon$ \textbf{then} \textbf{return} $\Delta_{\bm{\xi}}$.
\STATE \textbf{else} 
\STATE \quad $h^{(i)}$ $\leftarrow$ side length of the square to modify (according to some schedule \cite{andriushchenko2020square});
\STATE \quad $\Delta_{\bm{\xi}}^{\text{new}} \sim P\left(\rho, h^{(i)}, w, c, \Delta_{\bm{\xi}}, \hat{\x}, \x\right)$ for $\x \in \mathcal{D}_p$ (see \textbf{Appendix} \red{B} for details);
\STATE \quad $\hat{\x}_{\text {new }} \leftarrow \text { Project } \hat{\x}+\Delta_{\bm{\xi}}^{\text{new}} \text { onto }\left\{z \in \mathbb{R}^d:\|z-x\|_p \leq \rho\right\} \cap[0,1]^d$ for $\x \in \mathcal{D}_p$;
\STATE \quad $l_{\text {new }} \leftarrow \mathcal{L}_{tot}(\hat{\x}_{\text {new }},t)$ for $\x \in \mathcal{D}_p$; 
\STATE \quad $\text { if } l_{\text {new }}<l^* \text { then } \Delta_{\bm{\xi}} \leftarrow \Delta_{\bm{\xi}}^{\text{new}}, l^* \leftarrow l_{\text {new }}$, compute ASR;
\STATE \quad $i \leftarrow i+1$\text {; }
\STATE \textbf{end if} 

\ENDFOR
\RETURN $\Delta_{\bm{\xi}}^{*}$. 
\end{algorithmic}
\end{algorithm}

\section{Experiments\label{sec4}}

\subsection{Implementation details\label{sec4.1}}
\paragraph{Models and datasets.}
For image classification task, we evaluate all our attacks on three benchmark datasets CIFAR-10c\cite{krizhevsky2009learning}, Tiny ImageNet \cite{le2015tiny}, and GTSRB \cite{stallkamp2011german} over two network architectures, PreAct-ResNet18 \cite{he2016identity} and VGG19-BN \cite{simonyan2014very}. We utilize the implementation and setup in BackdoorBench \cite{wubackdoorbench}. For MMCL task, we use the open-sourced CLIP model from OpenAI \cite{radford2021learning} as the pre-trained model. Following the setting of CleanCLIP \cite{bansal2023cleanclip}, the model is poisoned on the CC3M dataset \cite{sharma2018conceptual} and subsequently tested through zero-shot evaluation on ImageNet-1K validation set \cite{deng2009imagenet}.
\paragraph{Backdoor attacks.}
For image classification task, we adopt seven widely used backdoor attacks including: (1) five data poisoning attack: BadNets \cite{gu2019badnets}, Blended \cite{chen2017targeted}, LF \cite{zeng2021rethinking}, SSBA \cite{li2021invisible}, and Trojan \cite{Trojannn}; and (2) two training-controllable attacks: Input-Aware \cite{nguyen2020input} and WaNet \cite{nguyen2021wanet}. We follow the default attack configuration as in BackdoorBench \cite{wubackdoorbench} and the $0^{th}$ label is set to be the target label. For MMCL task, we adopt four backdoor attacks including: BadNets, Blended, SIG \cite{barni2019new}, and TrojanVQA \cite{walmer2022dual}. In data poisoning phase, 1500 samples out of 500K image-text pairs from CC3M dataset are poisoned and the target label is \texttt{banana} as in \cite{bansal2023cleanclip}.

\paragraph{Backdoor defenses.}
For image classification task, we adopt six state-of-the-art post-training defense methods: NC \cite{wang2019neural}, NAD \cite{li2021neural}, i-BAU \cite{zeng2022adversarial}, FT-SAM \cite{zhu2023enhancing} , SAU \cite{wei2024shared}, and FST \cite{min2024towards}. For MMCL task, we consider two defense methods: (1) FT \cite{bansal2023cleanclip}: fine-tuning the model with multimodal contrastive loss using clean dataset; and (2) CleanCLIP \cite{bansal2023cleanclip}: a fine-tuning defense method for CLIP models.
 All the detailed introduction about the above attack and defense methods can be found in \textbf{Appendix} \red{B}.
\paragraph{Implementation details.}

At backdoor injection phase, the poisoning ratio is set to $10\%$, following the configuration in BackdoorBench \cite{wubackdoorbench}. At defense phase, $5\%$ clean samples are given to defend models. At backdoor re-activation phase, we consider defense models as our target model. The adversary is given $2\%$ (\ie, 1000) poisoned samples to conduct attacks. We consider both $\ell_{\infty}$ and $\ell_2$ norm attacks, and the perturbation bounds are set to 0.05 and 2, respectively. The loss hyper-parameter $\lambda$ is 1 for all our experiments. For query-based black-box attack, the maximum query limit is 10,000 for each image. For transfer attack, the adversary is given $10\%$ poisoned samples to conduct backdoor re-activation attack. The $\ell_2$ norm bound is set to 1 for transfer attack. We simply assume three surrogate models can be used and we just divided these defenses into two groups: (1) NC, NAD, i-BAU; and (2) FT-SAM, SAU, FST. Specifically, we generate perturbation in each group and test the ASRs in the other group. All the ASRs are tested on testing dataset. 
For MMCL tasks, we assume the adversary lacks knowledge of the downstream task. Therefore, attacks are executed in the upstream task for both white-box and transfer attacks and subsequently tested in downstream zero-shot task. More details about the implementations can be found in \textbf{Appendix} \red{D}.

\subsection{Main results\label{sec4.2}}

\paragraph{Backdoor re-activation attack.}
Tab. \ref{table1} shows the performance of our backdoor re-activation attack under white-box attack (\textbf{WBA}) and query-based black-box attack (\textbf{BBA}) settings in comparison with ASRs of original backdoored models (\textbf{No Defense}) and defense models (\textbf{Defense}). By observing the table, the following profound insights emerge: \textbf{(1)} Compared to defense models, our attacks show a striking level of efficacy. Both our WBA and BBA have exhibited an impressive improvement of 76.94\% and 42.95\% on average, respectively when compared against defense mechanisms, thereby underscoring security vulnerabilities in the defense models.
\textbf{(2)} The close performance of our WBA compared to "No Defense" underscores the efficacy of our backdoor re-activation mechanism, affirming the recoverability of the backdoors in defense models. By setting WBA as an upper bound for backdoor recovery, the more realistic BBA reveals substantial attack performance. Despite a gap between the two approaches, we posit that this disparity can be lessened through a sophisticated black-box attack strategy.
\textbf{(3)} In terms of specific defenses, our attack against SAU and FST exhibits relatively poor ASRs. This suggests that SAU's backdoor removal efficiency is significant, which aligns with the subsequent analysis of Fig. \ref{fig4.4}. In contrast, FST's BBA seems comparatively subdued. It may be attributed to the reinitialized FC layers in FST, effectively cutting backdoor activations. These insights serve as valuable pointers for crafting advanced defense strategies in the future.

\paragraph{Backdoor re-activation attack via transfer attack.} 
In this experiment, we groupe these defenses into two distinct groups: (1) weak group (NC, NAD, i-BAU) and (2) strong group (FT-SAM, SAU, FST) to better to observe the impact of defense methods on the performance of transfer-based re-activation attacks (\textbf{TA}). Two key findings emerged from results in Tab. \ref{table5}: \textbf{(1)} Transfer attacks generally exhibit strong performance in comparison with results in Tab. \ref{table1}. The ensemble attack strategies applied on weak group demonstrate better attack effectiveness on strong defense models than that in BBAs. \textbf{(2)} Utilizing ensemble strategies on strong defense methods results in remarkably effective ASRs on weak defense models, surpassing even the efficacy of WBA in Tab. \ref{table1}. This outcome raises concerns: if adversarys simulate stronger defenses to derive substitute models for launching transfer attacks, it could lead to serious security threats.

\begin{table}[t]
\caption{Performance (\%) of backdoor re-activation attack on both white-box (WBA) and black-box (BBA) scenarios with $\ell_{\infty}$-norm bound $\rho=0.05$ against different defenses with CIFAR-10 on PreAct-ResNet18. The best results are highlighted in \textbf{boldface}.}
\setlength{\tabcolsep}{2pt} 
\centering
\renewcommand\arraystretch{1.2}
\scalebox{0.58}{%
\begin{tabular}{c|c|ccc|ccc|ccc|ccc|ccc|ccc}
\toprule
\multirow{2}{*}{Attacks} &
  \multirow{2}{*}{No Defense} &
  \multicolumn{3}{c|}{NC \cite{wang2019neural}} &
  \multicolumn{3}{c|}{NAD \cite{li2021neural}} &
  \multicolumn{3}{c|}{i-BAU \cite{zeng2022adversarial}} &
  \multicolumn{3}{c|}{FT-SAM \cite{zhu2023enhancing}} &
  \multicolumn{3}{c|}{SAU \cite{wei2024shared}} &
  \multicolumn{3}{c}{FST  \cite{min2024towards}} \\
 &
   &
  Defense &
  WBA &
  BBA &
  Defense &
  WBA &
  BBA &
  Defense &
  WBA &
  BBA &
  Defense &
  WBA &
  BBA &
  Defense &
  WBA &
  BBA &
  Defense &
  WBA &
  BBA \\ \midrule
BadNets \cite{gu2019badnets} &
  93.79 &
  2.01 &
  \textbf{96.78} &
  27.91 &
  1.96 &
  \textbf{94.78} &
  49.66 &
  4.48 &
  \textbf{97.42} &
  54.37 &
  1.63 &
  \textbf{94.71} &
  51.23 &
  1.30 &
  \textbf{93.10} &
  37.91 &
  1.46 &
  \textbf{97.93} &
  42.69 \\
Blended \cite{chen2017targeted} &
  99.76 &
  99.76 &
  \textbf{99.93} &
  99.13 &
  47.64 &
  \textbf{99.82} &
  14.14 &
  26.83 &
  \textbf{99.63} &
  85.80 &
  12.17 &
  \textbf{99.56} &
  87.29 &
  5.20 &
  \textbf{98.37} &
  73.06 &
  0.20 &
  \textbf{99.62} &
  82.97 \\
Input-Aware \cite{nguyen2020input} &
  99.30 &
  0.70 &
  \textbf{92.04} &
  54.33 &
  0.92 &
  \textbf{93.80} &
  70.44 &
  0.02 &
  \textbf{21.78} &
  19.56 &
  1.07 &
  \textbf{96.19} &
  80.16 &
  1.26 &
  \textbf{85.39} &
  22.26 &
  0.00 &
  \textbf{90.72} &
  44.65 \\
LF \cite{zeng2021rethinking} &
  99.06 &
  99.06 &
  \textbf{99.41} &
  80.51 &
  75.47 &
  \textbf{99.41} &
  17.01 &
  11.99 &
  \textbf{99.04} &
  75.48 &
  6.43 &
  \textbf{97.40} &
  89.28 &
  2.49 &
  \textbf{90.74} &
  23.08 &
  5.43 &
  \textbf{98.18} &
  1.16 \\
SSBA \cite{li2021invisible} &
  97.07 &
  97.07 &
  \textbf{99.90} &
  94.38 &
  70.77 &
  \textbf{99.72} &
  88.53 &
  2.89 &
  \textbf{91.29} &
  70.71 &
  4.06 &
  \textbf{92.80} &
  69.18 &
  2.16 &
  \textbf{89.86} &
  38.59 &
  0.54 &
  \textbf{94.11} &
  52.71 \\
Trojan \cite{Trojannn} &
  99.99 &
  2.76 &
  \textbf{95.26} &
  45.57 &
  5.77 &
  \textbf{96.38} &
  60.87 &
  0.54 &
  \textbf{89.58} &
  40.18 &
  4.12 &
  \textbf{96.18} &
  69.88 &
  1.39 &
  \textbf{87.61} &
  47.37 &
  8.93 &
  \textbf{97.28} &
  80.47 \\
WaNet \cite{nguyen2021wanet} &
  98.90 &
  98.90 &
  \textbf{100.00} &
  99.64 &
  0.73 &
  \textbf{96.21} &
  77.65 &
  0.88 &
  \textbf{94.67} &
  75.91 &
  0.96 &
  \textbf{94.95} &
  78.66 &
  0.82 &
  \textbf{95.33} &
  60.36 &
  0.26 &
  \textbf{97.56} &
  82.22 \\ \midrule
Avg &
  98.26 &
  57.18 &
  \textbf{97.62} &
  71.64 &
  29.04 &
  \textbf{97.16} &
  54.04 &
  6.80 &
  \textbf{84.77} &
  60.29 &
  4.35 &
  \textbf{95.97} &
  75.10 &
  2.09 &
  \textbf{91.48} &
  43.23 &
  2.40 &
  \textbf{96.49} &
  55.27 \\ \bottomrule
\end{tabular}
\label{table1}
}
\end{table}
\begin{table}[t]
\caption{Attack performance (\%) on target models of transfer-based re-activation attack (TA) with $\ell_{2}$-norm bound $\rho=1$ against different defenses with CIFAR-10 on PreAct-ResNet18.}
\setlength{\tabcolsep}{3pt} 
\centering
\renewcommand\arraystretch{1.2}
\scalebox{0.6}{%
\begin{tabular}{c|c|cc|cc|cc|cc|cc|cc}
\toprule
\multirow{2}{*}{Attack} &
  \multirow{2}{*}{No Defense} &
  \multicolumn{2}{c|}{NC \cite{wang2019neural}} &
  \multicolumn{2}{c|}{NAD \cite{li2021neural}} &
  \multicolumn{2}{c|}{i-BAU \cite{zeng2022adversarial}} &
  \multicolumn{2}{c|}{FT-SAM \cite{zhu2023enhancing}} &
  \multicolumn{2}{c|}{SAU \cite{wei2024shared}} &
  \multicolumn{2}{c}{FST  \cite{min2024towards}} \\
                                   &       & Defense & TA     & Defense & TA    & Defense & TA    & Defense & TA    & Defense & TA    & Defense & TA    \\ \midrule
BadNets \cite{gu2019badnets}       & 93.79 & 2.01    & 95.43  & 1.96    & 98.42 & 4.48    & 97.90 & 1.63  & 97.42 & 1.30    & 90.17 & 1.46    & 96.21 \\
Blended \cite{chen2017targeted}    & 99.76 & 99.76   & 100.00 & 47.64   & 99.98 & 26.83   & 99.83 & 12.17   & 99.63 & 5.20    & 93.36 & 0.20    & 24.07 \\
Input-Aware \cite{nguyen2020input} & 99.30 & 0.70    & 99.98  & 0.92    & 99.98 & 0.02    & 99.77 & 1.07   & 21.78 & 1.26    & 15.56 & 0.00    & 95.28 \\
LF \cite{zeng2021rethinking}       & 99.06 & 99.06   & 99.93  & 75.47   & 99.84 & 11.99   & 98.35 & 6.43   & 99.04 & 2.49    & 96.62 & 5.43    & 80.09 \\
SSBA \cite{li2021invisible}        & 97.07 & 97.07   & 99.27  & 70.77   & 99.38 & 2.89    & 20.44 & 4.06   & 91.29 & 2.16    & 95.03 & 0.54    & 76.21 \\
Trojan \cite{Trojannn}             & 99.99 & 2.76    & 99.76  & 5.77    & 99.09 & 0.54    & 96.18 & 4.12   & 89.58 & 1.39    & 83.67 & 8.93    & 21.79 \\
WaNet \cite{nguyen2021wanet}       & 98.90 & 98.90   & 99.72  & 0.73    & 99.86 & 0.88    & 83.79 & 0.96   & 94.67 & 0.82    & 89.49 & 0.26    & 98.69 \\ \midrule
Avg                                & 98.26 & 57.18   & 99.16  & 29.04   & 99.51 & 6.80    & 85.18 & 4.35   & 84.77 & 2.09    & 80.55 & 2.40    & 70.33 \\ \bottomrule
\end{tabular}
\label{table5}
}
\end{table}

\begin{table}
\parbox{.59\linewidth}{
\caption{Performance (\%) of our attack on both white-box (WBA) and transfer-based (TA) attacks with $\ell_{\infty}$-norm bound $\rho=0.05$ against different defenses with ImageNet-1K on CLIP. Best results are highlighted in \textbf{boldface}.}\label{table6}
\setlength{\tabcolsep}{4pt} 
\renewcommand\arraystretch{1.2}
\scalebox{0.7}{%
\begin{tabular}{c|c|ccc|ccc}
\toprule
\multirow{2}{*}{Attack} & \multirow{2}{*}{No Defense} & \multicolumn{3}{c|}{FT \cite{bansal2023cleanclip}} & \multicolumn{3}{c}{CleanCLIP \cite{bansal2023cleanclip}} \\
                                &       & Defense & WBA            & TA             & Defense & WBA            & TA    \\ \midrule
BadNets \cite{gu2019badnets}    & 96.65 & 64.60   & 82.05          & \textbf{82.73} & 17.29   & \textbf{57.76} & 47.30 \\
Blended \cite{chen2017targeted} & 97.71 & 49.77   & 96.57          & \textbf{98.64} & 18.57   & \textbf{89.61} & 72.65 \\
SIG \cite{barni2019new}         & 77.71 & 30.91   & \textbf{92.56} & 87.99          & 21.68   & \textbf{87.04} & 82.55 \\
TrojanVQA \cite{walmer2022dual} & 98.21 & 82.07   & 97.14          & \textbf{97.46} & 49.82   & \textbf{87.43} & 78.25 \\ \midrule
Avg                             & 92.57 & 56.84   & \textbf{92.08} & 91.71          & 26.84   & \textbf{80.46} & 70.19 \\ \bottomrule

\end{tabular}
}}
\hspace{0.04cm}
\parbox{.39\linewidth}{
\centering
\caption{Our attacks (\%) on defense models in comparison with clean ones with $\ell_{\infty}$-norm bound $\rho=0.05$ under different model structures and datasets.} \label{table7}
\setlength{\tabcolsep}{4pt} 
\renewcommand\arraystretch{1.2}
\scalebox{0.7}{%
\begin{tabular}{c|cc|cc}
\toprule
\multirow{2}{*}{Setup} & \multicolumn{2}{c|}{Clean Model} & \multicolumn{2}{c}{Defense Model} \\
                       & WBA          & BBA          & WBA         & BBA        \\ \midrule
Res18+CIFAR-10          & 85.00        & 56.98        & 93.92       & 59.93      \\
Res18+Tiny             & 39.76        & 14.02        & 71.04       & 40.81      \\
Res18+GTSRB            & 53.33        & 50.87        & 67.14       & 61.81      \\
VGG+CIFAR-10            & 68.80        & 43.60        & 85.15       & 51.04      \\ \midrule
Avg                    & 61.72        & 41.37        & 79.31       & 53.40      \\ \bottomrule

\end{tabular}
}}

\end{table}

\paragraph{Effectiveness of attacks on CLIP models.}

Tab. \ref{table6} lists the performance of our backdoor re-activation attack under white-box attack (\textbf{WBA}) and transfer-based attack (\textbf{TA}) on the CLIP model. Our attacks yield significant improvements, with ASR enhancements of 34.87\% and 43.35\% on average, respectively, compared to defense models. The results for TA and WBA are very close. One possible reason is that the similarity between the FT and CleanCLIP methods leads to strong transfer performance. We advocate for the development of stronger defenses on CLIP to combat attacks.
Due to space constraints, attack results and analysis on Tiny ImageNet and GTSRB datasets, and results on VGG19-BN models are provided in \textbf{Appendix} \red{E}.

\begin{figure*}[htbp]
\centering
\subfigure{
\begin{minipage}[t]{0.5\linewidth}
\centering
\includegraphics[width=\textwidth]{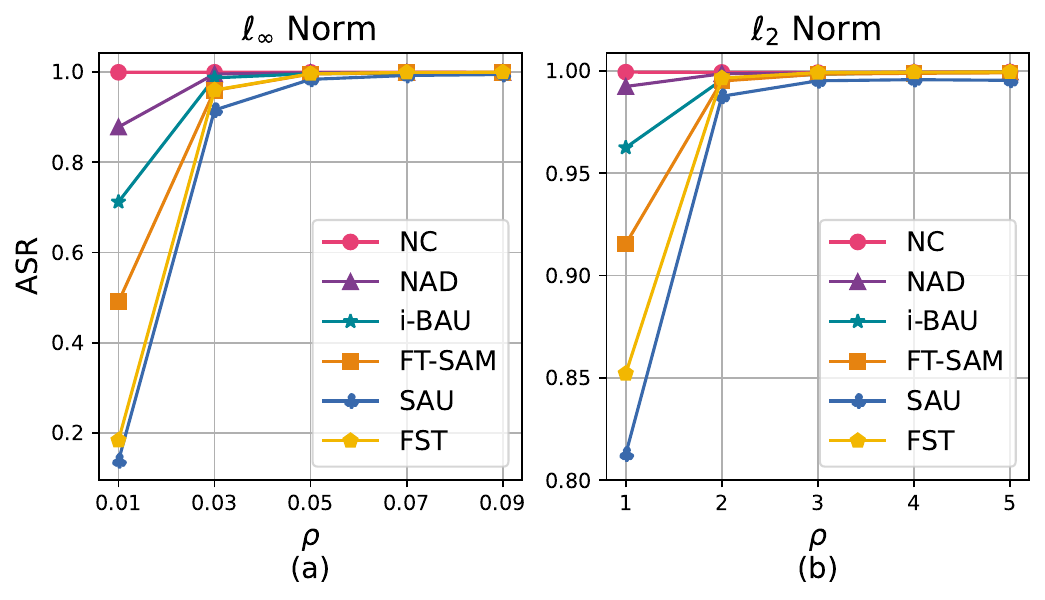}
\end{minipage}%
}%
\subfigure{
\begin{minipage}[t]{0.5\linewidth}
\centering
\includegraphics[width=\textwidth]{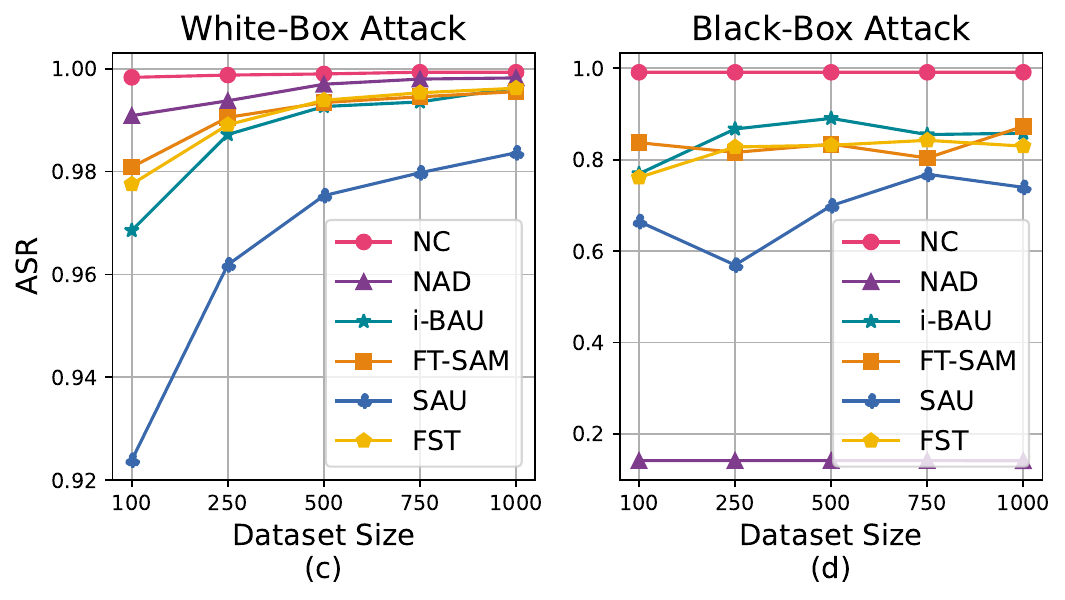}
\end{minipage}%
}%
\centering
\vspace{-4mm}
\caption{(a) and (b) show attack results under different norm types $p$ and bounds $\rho$ for WBA. (c) and (d) show attack results under different number of poisoned samples for WBA and BBA.}
\vspace{-4mm}
\label{fig:ablation}
\end{figure*}
\subsection{Ablation study\label{sec4.3}}
\paragraph{Influence of norm bound and norm type.} 
We studied the impact of norm type and norm bound on the attack performance. The results are shown in (a) and (b) of Fig. \ref{fig:ablation}. It can be observed that it is difficult to achieve high success rates under smaller norm bounds. However, when the norm bound is sufficiently large, the attack effectiveness converges and approaching nearly 100\% for both $\ell_{\infty}$-norm and $\ell_{2}$-norm types against all defense models.

\paragraph{Influence of the size of poisoned samples.}
We investigated the impact of the size of poisoned samples on attack performance for Blended attack. As shown in (c) and (d) of Fig. \ref{fig:ablation}, increasing the number of training samples in WBA shows significant improvement in attack results. However, in the BBA setting, the ASRs remains relatively stable and does not exhibit significant enhancements with the increase of training samples. This suggests that the difficulty in BBA lies in finding a good universal perturbation, especially when dealing with a large number of training samples.
However, the successful attacks with minimal samples also highlight the significant potency of the attack method.
\paragraph{Attacks performance against clean models.}

To demonstrate the specific vulnerability of defense models, we contrast the performance of our attacks on the defense models in comparison with clean models. Tab. \ref{table7} provides a summary of our method's performance across all backdoor attacks and defense methods, in comparison of the ASRs on clean models. It can be observed that, although some effectiveness is achieved on the clean models, the vulnerability of defense models is significantly higher than that of the clean model, with this gap being more pronounced in particular defenses. This indicates that defense models are indeed more fragile in comparison with clean models.

\subsection{Further analysis\label{sec4.4}}

\paragraph{Backdoor existence analysis.} 

\begin{figure*}
    \centering
    \includegraphics[width=1\linewidth]{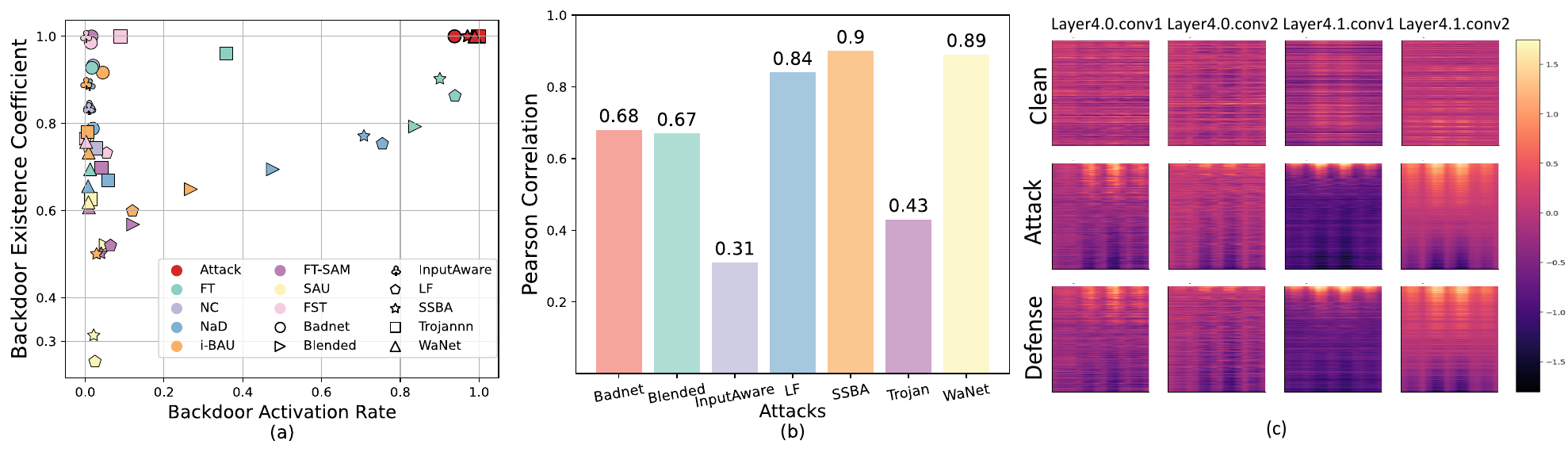}
    \caption{(a).Visualization of the correlation between backdoor activation rate and BEC. (b). Pearson correlation coefficients of ASR and BEC under different attacks. (c). Visualization of feature maps.}
    \label{fig4.4}
    \vspace{-4mm}
\end{figure*}

We provide more experimental demonstration on the existence of backdoors in defense models. We employ our BEC metric to quantify the existence of backdoors in all defense models and visualize the relationship between BEC and backdoor activation rates, as depicted in (a) of Fig. \ref{fig4.4}. We observe that backdoors persist across defense models, albeit with low backdoor activation rates. The BECs in SAU, SAM, and i-BAU are relatively low, while FST exhibits a notably high BEC. This contrast may stem from the former's optimization objectives resembling adversarial training, whereas the latter primarily disrupts activations through layers re-initialization.
\paragraph{Relationship between BECs and ASRs.} 

We validate the relationship between the ASR of re-activation attack (WBA) and the residual of backdoors. We computed the Pearson Correlation Coefficients (PPC) between BECs of different defense models and their white-box ASRs among all attacks, as shown in (b) of Fig. \ref{fig4.4}. It is evident that in most cases, there is a strong correlation between the two. In other words, the more backdoors remain in models, the easier it is for attacks to succeed. Therefore, our metrics can serve as an indicator of backdoored model security.

\paragraph{Feature map visualization.} 

Here we visualize the feature maps between different models to directly observe their similarities. Fig. \ref{fig4.4} (c) displays the visualizations of activations from the final four convolutional layers of three models, sorted in descending order according to backdoored model's TAC value, with each subplot arranged from top to bottom. It can be observed that the defense model and backdoored model exhibit similar patterns: highlighting activations in backdoor-related neurons. This directly indicates the persistence of backdoors within defense models.

\begin{wraptable}{R}{7cm}
\vspace{-0.2in}
\caption{Results (\%) against adaptive defense.}\label{table8}
\setlength{\tabcolsep}{3pt} 
\renewcommand\arraystretch{1.2}
\scalebox{0.7}{
\begin{tabular}{c|cc|cc|cc|cc}
\toprule
Defense $\rightarrow$ &
  \multicolumn{2}{c|}{FST  \cite{min2024towards}} &
  \multicolumn{2}{c|}{FT-SAM \cite{zhu2023enhancing}} &
  \multicolumn{2}{c|}{i-BAU \cite{zeng2022adversarial}} &
  \multicolumn{2}{c}{SAU \cite{wei2024shared}} \\
Noise $\downarrow$ & ACC   & ASR   & ACC   & ASR   & ACC   & ASR   & ACC   & ASR   \\ \midrule
0.00  & 92.61 & 82.97 & 92.88 & 87.29 & 89.43 & 85.80 & 91.75 & 73.06 \\
0.01  & 91.64 & 82.24 & 92.13 & 90.89 & 88.85 & 77.88 & 91.31 & 71.99 \\
0.02  & 88.35 & 70.89 & 89.53 & 88.04 & 86.15 & 79.02 & 88.38 & 84.11 \\
0.03  & 82.59 & 73.03 & 84.84 & 86.36 & 81.32 & 75.83 & 83.33 & 67.42 \\
0.04  & 75.87 & 56.76 & 78.04 & 81.60 & 75.51 & 72.72 & 76.19 & 54.40 \\
0.05  & 67.15 & 58.17 & 70.12 & 74.57 & 68.09 & 72.28 & 67.95 & 56.74 \\ \bottomrule
\end{tabular}}
\end{wraptable}

\paragraph{Attack against adaptive defense.} 

Considering defenders are aware of adversary' strategies, they can introduce random perturbations for each query so as to disrupt the adversary's ability. We assess both adversary's ASR and the model accuracy on clean samples under varying perturbation bound. As depicted in Tab. \ref{table8}, minor noise has slight impact on ASR. However, with larger noise amplitudes, despite failed attacks, the model's accuracy is significantly affected.

\section{Conclusion}
This paper illuminates the false sense of security in backdoor defenses and proposes a new threat to enhance existing backdoor attacks in inference-time. Our pioneering introduction of the backdoor existence coefficient unveils the residual presence of backdoors within defense models. Moreover, we propose a novel optimization problem to re-activate these dormant backdoors and craft distinct algorithms tailored specifically to white-box, black-box, and transfer attack scenarios. The proposed method can be integrated with existing backdoor attacks to boost their attack success rate during the inference stage. The efficacy of our method is evidenced through exhaustive evaluation on both image classification and multi-modal contrastive learning tasks. The threat revealed by this study underscores the pressing need for  designing advanced defense mechanisms in the future.

\paragraph{Limitations and future work.}
Despite the efficacy of our proposed method, its effectiveness is limited when confronted with defenders that inject noise into each query. Promising future work is to devise more sophisticated attacks that can bypass this defenses. Another limitation is that if defenders aim to decrease both ASR and BEC, our attacks will become challenging, even though directly optimizing the BEC is not feasible. This serves as another direction for our future work.

\paragraph{Broader Impacts.}

As deep neural networks sourced from untrusted origins face significant risks from backdoor attacks, this study provides a meaningful exploration into the false security in backdoor defense models. This could spark further advancements in backdoor defenses. Nonetheless, the potential misuse by ill-intended entities should be cautiously considered.

\clearpage
{\small
\bibliography{main}
}
\end{document}